\documentclass[sigconf]{acmart}
\preto{\abstractkeywords}{\nolinenumbers} 

\usepackage{subfig}
\usepackage{colortbl} 
\usepackage{xcolor}   
\usepackage{hyperref}
\usepackage{url}
\usepackage{lineno}

\AtBeginDocument{%
  }

\setcopyright{acmlicensed}
\renewcommand\footnotetextcopyrightpermission[1]{}
\copyrightyear{2018}
\acmYear{2018}
\acmDOI{XXXXXXX.XXXXXXX}
\acmConference{}{}{}
\author{Jiahuan Long}
\affiliation{
  \institution{Chinese Academy of Military Science}
  \city{Beijing} \country{China}
}

\author{Wen Yao}
\affiliation{
  \institution{Chinese Academy of Military Science}
  \city{Beijing} \country{China}
}

\author{Tingsong Jiang}
\affiliation{
  \institution{Chinese Academy of Military Science}
  \city{Beijing} \country{China}
}

\author{Chao Ma}
\affiliation{
  \institution{Shanghai Jiao Tong University}
  \city{Shanghai} \country{China}
}

\acmSubmissionID{2511}

\begin{document}

\title{CDUPatch: Color-Driven Universal Adversarial Patch Attack for Dual-Modal Visible-Infrared Detectors}

\begin{abstract}
Adversarial patches are widely used to evaluate the robustness of object detection systems in real-world scenarios. These patches were initially designed to deceive single-modal detectors (e.g., visible or infrared) and have recently been extended to target visible-infrared dual-modal detectors. However, existing dual-modal adversarial patch attacks have limited attack effectiveness across diverse physical scenarios. To address this, we propose CDUPatch, a universal cross-modal patch attack against visible-infrared object detectors across scales, views, and scenarios. Specifically, we observe that color variations lead to different levels of thermal absorption, resulting in temperature differences in infrared imaging. Leveraging this property, we propose an RGB-to-infrared adapter that maps RGB patches to infrared patches, enabling unified optimization of cross-modal patches. By learning an optimal color distribution on the adversarial patch, we can manipulate its thermal response and generate an adversarial infrared texture. Additionally, we introduce a multi-scale clipping strategy and construct a new visible-infrared dataset, MSDrone, which contains aerial vehicle images in varying scales and perspectives. These data augmentation strategies enhance the robustness of our patch in real-world conditions.
Experiments show that our method outperforms existing attacks in the digital domain and demonstrates strong transferability across scales, views, and scenarios in physical tests.\end{abstract}

\begin{teaserfigure}
\includegraphics[width=\textwidth]{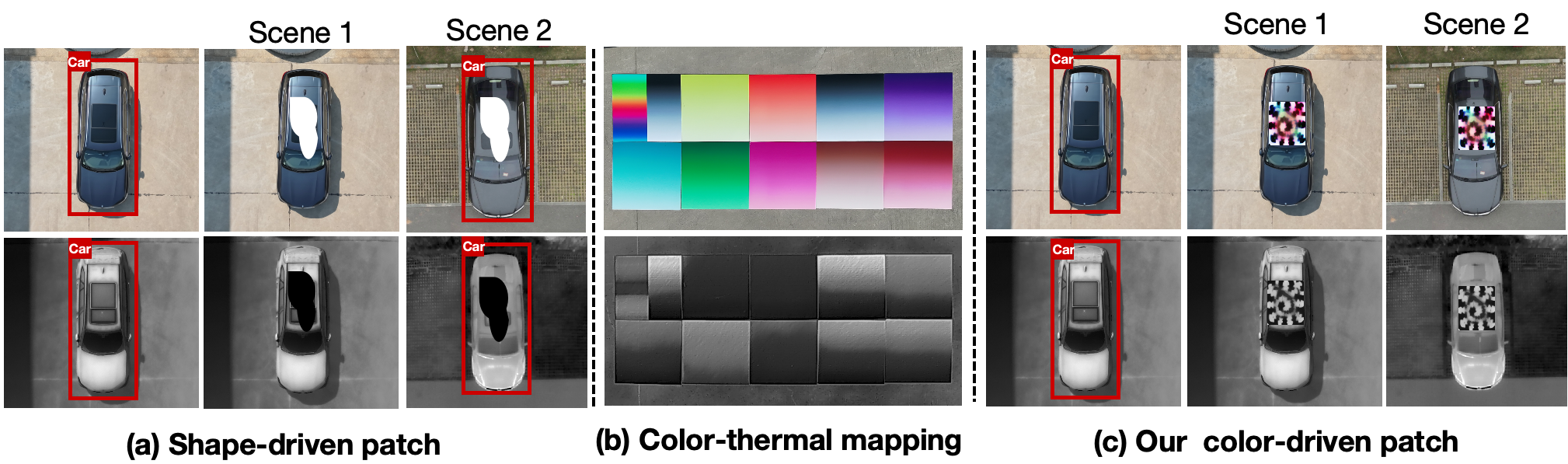}
  \caption{ Comparison between shape-driven and color-driven adversarial patches in visible and infrared modalities.  
(a) Shape-driven patch
effectively deceives the dual-modal detector but exhibits poor transferability across scenes.  
(b) Foam boards of different colors exhibit distinct thermal responses under infrared imaging.  
(c) Leveraging the phenomenon observed in (b), the proposed color-driven patch achieves effective dual-modal attacks and maintains strong generalization across diverse scenes.}
  \label{fig:teaser}
  \Description{Placeholder description.}
\end{teaserfigure}

\maketitle
\section{Introduction}
Dual-modal object detection (RGB + Infrared) is increasingly used in applications like intelligent surveillance~\cite{cheng2023towards, hou2024object} and drone-based search and rescue~\cite{ding2021object}. Infrared cameras offer stable thermal cues in challenging conditions, while RGB cameras capture detailed textures. Combining both enables reliable object identification across varied environments. However, as dual-modal detectors improve, their vulnerability to physical adversarial attacks, such as adversarial patches~\cite{thys2019fooling, TSEA, long2024papmot, zhu2023tpatch, hu2022adversarial}, has gained attention. These attacks involve physical perturbations that deceive detectors, compromising model robustness and posing safety risks. Thus, developing physical patch attacks for dual-modal detectors is essential to understanding their vulnerabilities.

Existing dual-modal adversarial patch methods are mainly shape-driven~\cite{tpamipatch, wei2023iccv, hupatch}. These approaches optimize a fixed geometric shape and then apply thermal reflective materials to achieve unified texture in visible and infrared domains. 
However, these shape-driven methods are typically optimized on a single visible-infrared image pair, resulting in locally optimal patch shapes that fail to generalize across different objects or scenarios in the real world.
As shown in 
Figure~\ref{fig:teaser}(a), the shape-driven patch optimized for one vehicle in Scene 1 fails to attack a different vehicle in Scene 2. 

To pursue a more generalized solution, we aim to design a universal cross-modal adversarial patch that remains effective across various scenes, targets, and viewpoints. However, achieving this goal involves two key challenges.  First, existing universal patch attacks~\cite{TSEA, thys2019fooling, hu2022adversarial, tang2023adversarial} are optimized only in the visible domain, limiting their effectiveness in the infrared domain. Thus, how to construct a unified optimization framework that bridges the visible and infrared modalities is the first challenge. Second, most aerial vehicle datasets~\cite{visdrone, dronevehicle} have limited scale diversity, restricting the attack transferability of patches. Thus, how to expand datasets to train a robustness patch across multi-scale scenarios is another challenge. 

To address these challenges, we observe that different colors exhibit varying infrared responses due to differences in thermal absorption. As shown in Figure~\ref{fig:teaser}(b), we printed color-gradient foam boards and captured them in both visible and infrared images. The results reveal that different colors produce distinct thermal patterns, e.g., green absorbing more heat than red, and black absorbing more heat than white. 
Building on this insight, we aim to optimize the color distribution on the adversarial patch to induce infrared textures with adversarial effects in the infrared domain. To achieve this, we design a color-driven cross-modal adversarial patch framework that leverages color-thermal differences. It incorporates a learnable RGB-to-infrared adapter that maps RGB patch into corresponding infrared textures. 
This adapter enables joint optimization across both modalities by allowing gradients to propagate from infrared predictions back to the RGB space.
To enhance patch robustness across object scales and viewpoints, we propose a multi-scale clipping strategy to augment existing datasets. Additionally, we collect a new dataset, MSDrone, consisting of visible-infrared aerial vehicle images under diverse scales, perspectives, and environments. These data augmentation strategies significantly improve the real-world effectiveness of our patch. As shown in Figure~\ref{fig:teaser}(c), our patch achieves consistent attack performance across varied scenes.

Our main contributions are summarized as follows:
\begin{itemize}
    \item We deeply explore physical properties of color–thermal imaging in the real world, and propose a universal color-driven adversarial patch optimization framework. The optimized patch can fools visible-infrared object detectors across different scenarios.
    
    \item We introduce a learnable RGB-to-infrared adapter that maps RGB patches to simulated infrared patches, which enables joint training of the adversarial patch across both modalities.

    \item We introduce
    a multi-scale clipping strategy and release a visible-infrared dataset, MSDrone, which helps to extend existing datasets and facilitates robust patch training across scales.
    
    \item Experiments on four datasets show that our method outperforms baselines in the digital domain, while extensive physical tests demonstrate strong attack transferability across scales, views, and scenarios.

\end{itemize}

\section{Related Work}
{\flushleft\bf Visible-Infrared Dual-Modal Detection}
Dual-modal detection systems~\cite{zhang2023differential, wang2023interactively}, which combine visible (RGB) and infrared (IR) modalities, are crucial for applications such as intelligent surveillance and border patrol. These systems are typically trained on datasets such as DroneVehicle~\cite{dronevehicle} and LLVIP~\cite{LLVIP}, which provide paired RGB and infrared images of vehicles and pedestrians. However, most existing dual-modal datasets lack sufficient scale diversity in vehicle objects, which limits the effectiveness of cross-modal adversarial patch training.
For example, the DroneVehicle dataset primarily features similarly sized vehicles and densely packed scenes, limiting the generalization of patches across object scales and complex environments. In this paper, we enhance training diversity by applying a multi-scale clipping strategy to augment existing datasets. In addition, we introduce MSDrone, a new dual-modal vehicle dataset containing RGB-IR image pairs captured under diverse object scales, viewpoints, and environmental conditions. Together, these improvements facilitate the training of more robust adversarial patches that generalize effectively across a wide range of real-world scenarios.

\begin{table}[t]
\centering
\vspace{-0.1in}
\caption{Comparison of representative works on adversarial attacks.}
\resizebox{0.47\textwidth}{!}{
\begin{tabular}{lcccccc}
\toprule
Methods & \cite{tpamipatch} & \cite{weiinfra} & \cite{thys2019fooling} & \cite{hupatch} & \cite{wu2025gradient} & Ours \\
\midrule
Attack transferability &     &     & $\surd$     &     &     &$\surd$     \\
Physical deployable   &$\surd$     &$\surd$     &$\surd$     &$\surd$     &     &$\surd$     \\
Implement difficulty       &Medium     &Medium     &Easy     &Hard     &Hard     &Easy     \\
Infrared effectiveness      &$\surd$     &$\surd$     &     &$\surd$     &     & $\surd$    \\
Visible effectiveness       &$\surd$     &     &$\surd$     &$\surd$     &$\surd$     &$\surd$     \\
Data augmentation       &     &     &     &     &     &$\surd$     \\
\bottomrule
\end{tabular}
}
\label{tab:different patch attacks}
\end{table}

{\flushleft\bf Adversarial Patch}
Previous digital attacks~\cite{xu2020adversarial1, jia2020robust, jia2021iou, jia2022exploring, long2025robust} are difficult to deploy in physical scenarios, as their perturbations are often imperceptible to the human eye and hard to reproduce in the real world. In contrast, adversarial patches have emerged as one of the most widely adopted physical attack methods against AI systems~\cite{thys2019fooling, TSEA, long2024papmot, zhu2023tpatch, hu2022adversarial, brown2017adversarial, liu2018dpatch}. These methods can be categorized into two main types: shape-driven and color-driven adversarial patches. Shape-driven patches~\cite{wei2023iccv, tpamipatch, hupatch} aim to optimize a fixed attack shape to mislead detectors. For instance, Wei et al.~\cite{tpamipatch} propose a novel boundary-limited shape optimization approach that aims to achieve compact and smooth shapes for the adversarial patch. However, such methods are typically trained on a single visible-infrared image pair, leading to overfitting and poor generalization across different objects or environments.
In contrast, color-driven patches~\cite{thys2019fooling, TSEA, long2024papmot} aim to optimize the color distribution of adversarial patches with strong attack transferability. For example, Huang et al.~\cite{TSEA} generate transferable adversarial patches for pedestrian detection and achieve a high ASR (72\%) on the INRIA dataset~\cite{inria}.  
While these methods generalize better across scenes, they are designed solely for the visible modality and fail to attack in the infrared domain. 

In this paper, we propose CDUPatch, the color-driven universal optimization framework to generate a cross-modal patch against visible-infrared detectors.  CDUPatch effectively attacks both visible and infrared modalities across diverse datasets (e.g., DroneVehicle~\cite{dronevehicle}, LLVIP~\cite{LLVIP}, VisDrone~\cite{visdrone}, and MSDrone). It is easy to deploy physically using a single foam board and demonstrates strong attack transferability across scenes, viewpoints, and object scales in the real world. 
A comparison of the features and contributions of our method against prior works is provided in Table~\ref{tab:different patch attacks}.

\begin{figure*}[t]
\centering
\includegraphics[scale=0.39]{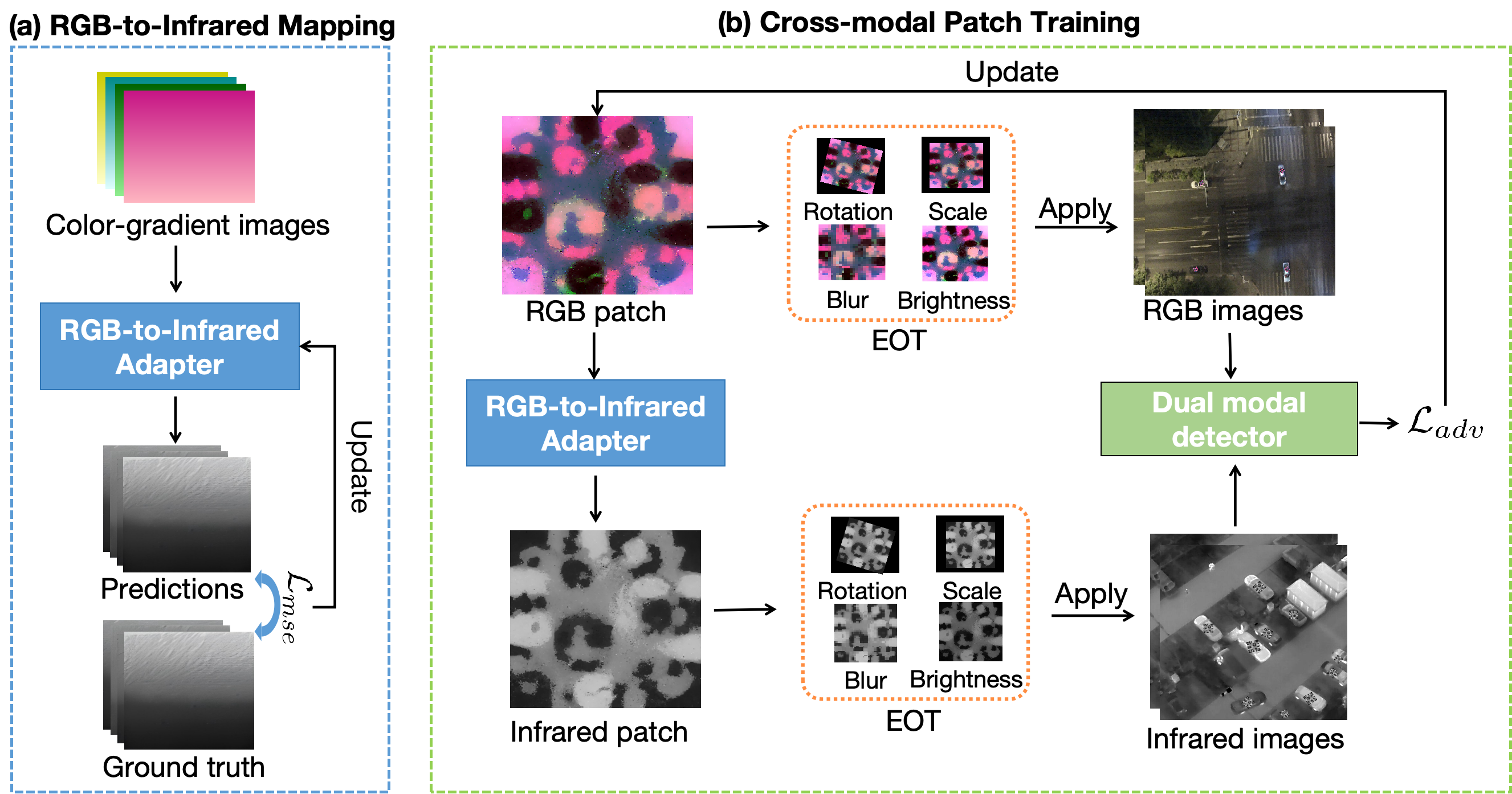}
\caption{The overview of CDUPatch. (a) represents RGB-to-infrared adapter training. It uses printed color-gradient images to learn the mapping from visible to infrared textures. (b) shows the cross-modal patch training. The RGB patch is directly applied to visible images, while its infrared counterpart is generated through a trained RGB-to-infrared adapter and then applied to infrared images. During this phase, Expectation over Transformation (EOT) is applied to ensure patch robustness under diverse physical transformations.}
\label{fig:framework}
\Description{Placeholder description.}
\end{figure*}

\section{Methodology}

\subsection{Preliminaries}
{\flushleft\bf Color-Thermal Effect in the Physical World.} Under solar radiation, an object’s surface temperature is primarily influenced by its material and surface color. According to the Stefan–Boltzmann Law~\cite{howell2020thermal}, this relationship can be described by the thermal equilibrium model:
\begin{equation}
T_{\text{surface}} = \left( \frac{\alpha G}{\varepsilon \sigma} + T_{\text{ambient}}^4 \right)^{1/4},
\end{equation}
where $\alpha$ is the color-dependent solar absorptivity, $G$ is solar irradiance, $\varepsilon$ is the material's emissivity, $\sigma$ is the Stefan–Boltzmann constant, and $T_{\text{ambient}}$ is the ambient temperature. For instance, black surfaces ($\alpha \approx 0.9$) absorb far more heat than white surfaces ($\alpha \approx 0.2$), leading to significant thermal contrast.

In our implementation, patches are printed on foam boards with varying color patterns. 
Because colors absorb heat differently, colored regions on the foam board form non-uniform temperatures under solar radiation, which in turn creates distinct textures in infrared imaging. 
Notably, solar radiation is always present under different weather conditions (e.g., sunny or cloudy), making the color-driven thermal textures consistently observable, as illustrated by the representative cases in Figure~\ref{fig:RGB-infrared validation}.

\subsection{Attack Formulation}

Let $\mathcal{D} = \{(I^v_j, I^r_j)\}_{j=1}^{N}$ be a dataset of dual-modal image pairs, where each sample consists of a visible image $I^v_j$ and its aligned infrared image $I^r_j$. Our objective is to optimize a single adversarial patch $p$ that, when simultaneously applied to both $I^v_j$ and $I^r_j$, can fool the visible-infrared detector.

We define a binary mask $\mathcal{A}_j$ that controls the location, size, and shape of the patch in image pair $j$. To account for real-world transformations (e.g., rotation, scaling), we adopt the Expectation over Transformation (EOT) framework and define the patch-injected images as:
\begin{equation}
    \tilde{I}^v_j = (1 - \mathcal{A}_j) \odot I^v_j + \mathcal{A}_j \odot E(p), \quad
\end{equation}
\begin{equation}
    \tilde{I}^r_j = (1 - \mathcal{A}_j) \odot I^r_j + \mathcal{A}_j \odot E(\mathcal{F}(p)),
\end{equation}
where $E(\cdot)$ denotes a differentiable transformation sampling function, $p$ is the learnable RGB patch, and $\mathcal{F}$ is a learnable mapping function that translates $p$ into its infrared patch.

Let $T$ denote the visible-infrared object detectors with parameters $\theta$. The patch is optimized to jointly reduce detection confidence in both visible and infrared domains:
\begin{equation}
p = \arg \min_{\kappa} \mathcal{L}_{adv}(T(\tilde{I}^v_j, \tilde{I}^r_j|\theta), \mathbb{D}_j, \kappa),
\end{equation}
where $\kappa$ denotes the pixel values of $p$, $\mathbb{D}_j$ is the set of detection outputs (or ground truth) for image pair $j$, and $\mathcal{L}_{adv}$ is the adversarial loss function, further described in Sec.~3.6.

\begin{figure}[t]
\centering
\includegraphics[scale=0.33]{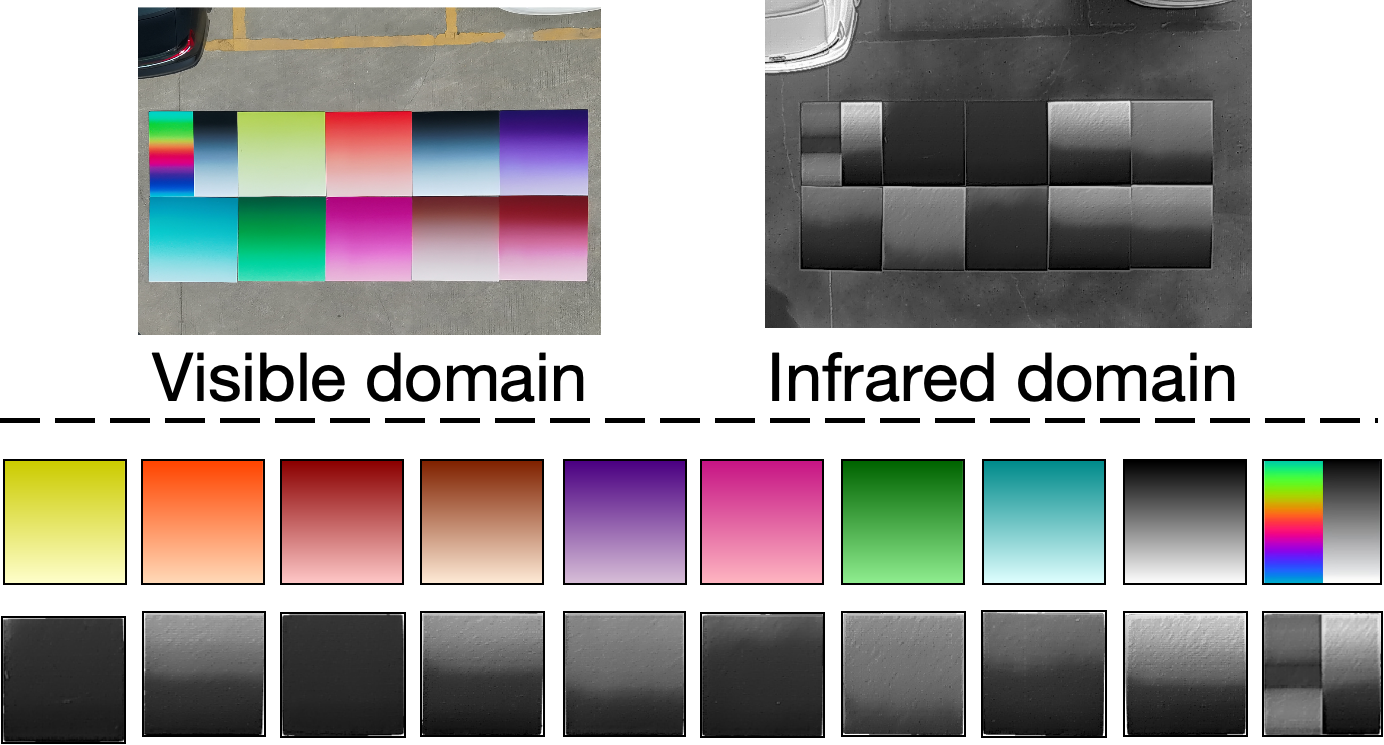}
\caption{Visible-infrared image pairs of printed color-gradient foam boards. These paired images are used to train our RGB-to-infrared adapter. The printed color patterns span diverse hues, providing rich chromatic diversity.}
\label{fig:color-gradient}
\Description{Placeholder description.}
\end{figure}

\subsection{Overall Framework of CDUPatch}
As mentioned in Section 3.1, different colors absorb solar radiation unevenly, leading to non-uniform temperature distributions and consequently forming distinct infrared textures.
The core objective of CDUPatch is to optimize a colored adversarial patch whose visible texture and infrared texture both exhibit strong and consistent attack effectiveness. The overall framework comprises two key components: RGB-to-Infrared Mapping and Cross-Modal Patch Training, as depicted in
Figure~\ref{fig:framework}.

As shown in Figure~\ref{fig:framework}(a), we introduce a learnable RGB-to-Infrared adaptor to bridge the patch appearance gap between visible and infrared domains. We print colorful gradient patterns on foam boards and capture them using RGB and infrared cameras at the same time (Figure~\ref{fig:color-gradient}). This gives us aligned image pairs, where each RGB pixel has a matching grayscale infrared value. 
The adapter is supervised using a mean squared error (MSE) loss $\mathcal{L}_{\text{mse}}$, which enforces pixel-wise similarity between the predicted results and ground-truth infrared data.
Once trained, this adapter is used to translate visible patches into infrared patches, enabling joint optimization across both modalities.


As illustrated in Figure~\ref{fig:framework}(b), we perform cross-modal adversarial patch training based on a two-branch structure. One branch operates on visible images, while the other processes infrared images. Starting from an initialized RGB patch, we apply it to both the visible and infrared branches. For the infrared branch, the RGB patch is first translated into an infrared patch via the RGB-to-Infrared adapter, and then applied to infrared images. For the visible branch, the original RGB patch is directly pasted onto RGB images. In the patch pasting process, we adopt the Expectation Over Transformation (EOT)~\cite{athalye2018synthesizing} to simulate real-world transformations (e.g., rotation, brightness changes, blur, and scaling), thereby improving the patch's robustness under diverse physical conditions. 
Finally, the patched images from both branches are forwarded to a dual-modal detector (e.g., YOLOv8~\cite{yolov8}), and the patch is updated via adversarial loss to suppress detection confidence in both modalities.

\begin{figure}[t]
\centering
\includegraphics[scale=0.25]{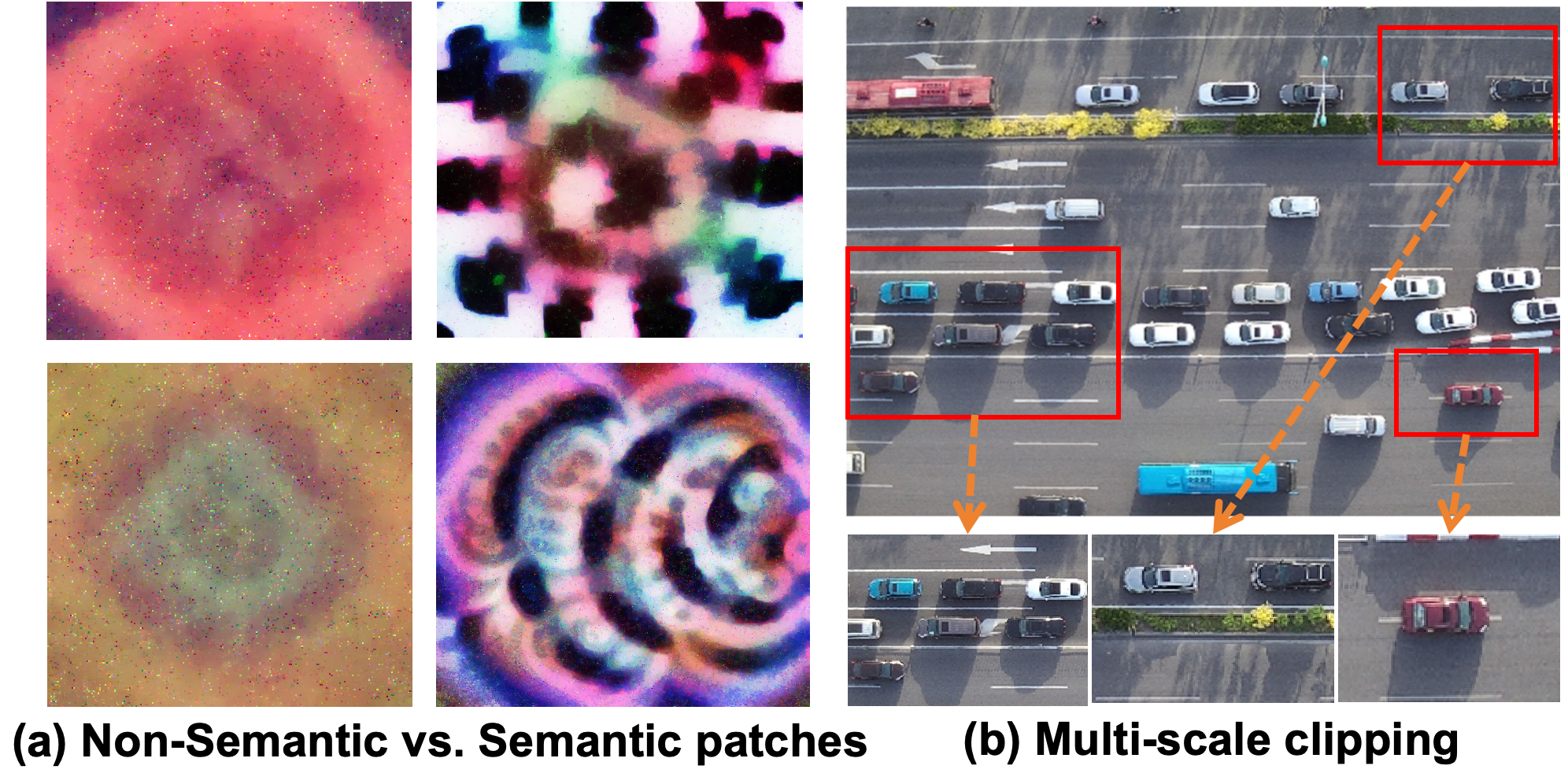}
\caption{Data augmentation for patch training. (a) represents optimized patches without (left) and with (right) semantic information. The latter has higher attack transferability. (b) shows the clip strategy that crops vehicle regions at multiple scales from high-resolution aerial images to enrich object size diversity. }
\label{fig:augmentation}
\Description{Placeholder description.}
\end{figure}

\subsection{RGB-to-Infrared Adapter}
To bridge the appearance gap between visible and infrared modalities, we design a lightweight visible-to-infrared adapter that learns a pixel-wise mapping from RGB color space to grayscale infrared intensity. To train this adapter, 
we print a series of color-gradient patterns on foam boards and capture them using synchronized visible and infrared cameras under stable lighting conditions. As shown in Figure~\ref{fig:color-gradient}, the printed color patterns span a wide range of hues and intensities, providing sufficient chromatic diversity for training the spectral mapping.

We then use the resulting paired RGB-infrared dataset to train a lightweight three-layer multilayer perceptron (MLP) as the RGB-to-infrared adapter. The network takes RGB pixel values $x \in \mathbb{R}^3$ as input and predicts the corresponding infrared grayscale intensity $\hat{y} \in \mathbb{R}$.
The model is optimized using the mean squared error (MSE):
\begin{equation}
\mathcal{L}_{mse} = \frac{1}{N} \sum_{i=1}^{N} \left( f(x_i) - y_i \right)^2,
\end{equation}
where $f(\cdot)$ denotes the adapter network, and $N$ is the number of sampled pixels.
By training this RGB-to-infrared adapter, we effectively capture the spectral mapping between RGB and infrared domains, enabling reliable translation of visible patches into infrared-style textures for cross-modal attack optimization.

\subsection{Data Augmentation}
Existing datasets such as DroneVehicle~\cite{dronevehicle} and VisDrone~\cite{visdrone} exhibit limited object scale variation, as most vehicles are captured from fixed altitudes and views.  As a result, adversarial patches trained on these datasets often lack semantic structure and show poor generalization to diverse object sizes and scenes. Prior work has shown that semantically meaningful patches are known to yield better transferability~\cite{long2024papmot,TSEA, tang2023adversarial}. As illustrated in Figure~\ref{fig:augmentation}(a), patches trained without semantic information tend to produce unstructured and visually meaningless patterns, whereas semantically enriched patches yield more coherent and transferable adversarial textures.

To address this, we adopt two data augmentation strategies to enrich scale and scene diversity in training data.
First, we propose a multi-scale clipping strategy, which augments existing aerial datasets by cropping vehicles at different resolutions and locations (see Figure~\ref{fig:augmentation}(b)). This encourages the patch to learn scale-variant adversarial patterns.
Second, we collect a new dataset named MSDrone, which contains over 2,000 paired visible-infrared images captured at varying distances and viewpoints.  This enhances scene diversity and supports robust training across both modalities. Details about the MSDrone dataset are provided in the Appendix.

\subsection{Adversarial Losses}
For our loss formulation, we adopt the widely used average precision loss and total variation loss~\cite{long2024papmot, Sharif_Bhagavatula_Bauer_Reiter_2016, hu2021naturalistic, FCA} to reduce the confidence score of detection while encouraging visual smoothness of the printable patches. 
{\flushleft\bf Total Variation Loss ($\mathcal{L}_{tv}$).} The aim of $\mathcal{L}_{tv}$ is to ensure the generation of printable patches with smooth color transitions, which helps to reduce the presence of visual noise. $\mathcal{L}_{tv}$ is formulated as:
\begin{equation}
\mathcal{L}_{t v}=\sum_{i, j} \sqrt{\left(\left(I_{i, j}-I_{i+1, j}\right)^{2}+\left(I_{i, j}-I_{i, j+1}\right)^{2}\right.},
\end{equation}
where $I_{i, j}$ denotes the pixel value at position ${(i, j)}$ in the patch. 

{\flushleft\bf Average Precision Loss ($\mathcal{L}_{ap}$).} 
The aim of $\mathcal{L}_{ap}$ is to reduce the average detection score of targets that have been attached to an adversarial patch. Given a set of $N$ detection boxes, where the detection score for the $i$-th box is $s_i$, $\mathcal{L}_{ap}$ is calculated as  follows:
\begin{equation}
\mathcal{L}_{ap} = \frac{1}{N} \sum_{i=1}^{N} s_i.
\end{equation}
Finally, the overall adversarial loss $L_{adv}$ is a sum of the above two loss terms weighted by hyperparameters  $\gamma$ and $\delta$:
    $\mathcal{L}_{adv}=\gamma \mathcal{L}_{tv}+\delta \mathcal{L}_{ap}$.

\section{Experiments}
\subsection{Experimental Setup}
{\flushleft\bf Datasets and Metrics.} 
We evaluate our attack method on three vehicle object detection benchmarks: DroneVehicle~\cite{dronevehicle},  VisDrone~\cite{visdrone}, LLVIP~\cite{LLVIP} and our custom multi-scale dataset, MSDrone. DroneVehicle~\cite{dronevehicle}, VisDrone~\cite{visdrone}, and LLVIP~\cite{LLVIP} are publicly available visible-infrared datasets with diverse vehicle objects. In addition, we introduce the MSDrone dataset that includes vehicle annotations across varying sizes and viewpoints, enabling a more comprehensive evaluation under scale variation.
For metrics, we adopt the \textit{Attack Success Rate} (ASR) as the primary evaluation metric, which quantifies the proportion of successfully suppressed detections under adversarial patch attack. We set the detection confidence threshold to 0.5 in all experiments. ASR is defined as:
\begin{equation}
\text{ASR} = \frac{N_{\text{clean}} - N_{\text{patch}}}{N_{\text{clean}}},
\end{equation}
where $N_{\text{clean}}$ and $N_{\text{patch}}$ denote the number of detected objects on clean and adversarial images, respectively.

\begin{table*}[t]
\centering
\caption{Comparison of different patch attacks across datasets and models (Attack Success Rate \%). }
\vspace{-0.05in}
\resizebox{0.98\textwidth}{!}{
\begin{tabular}{lcccccccccccccccc}
\toprule
\textbf{Methods} & \multicolumn{4}{c}{DroneVehicle~\cite{dronevehicle}} & \multicolumn{4}{c}{VisDrone~\cite{visdrone}} & \multicolumn{4}{c}{LLVIP~\cite{LLVIP}} & \multicolumn{4}{c}{MSDrone} \\
\cmidrule(lr){2-5} \cmidrule(lr){6-9} \cmidrule(lr){10-13} \cmidrule(lr){14-17}
 & v3 & v5 & v8 & RCNN & v3 & v5 & v8 & RCNN & v3 & v5 & v8 & RCNN & v3 & v5 & v8 & RCNN \\
\midrule
RandomPatch        
& 11.2 & 8.8 & 2.4 & 8.3
& 13.7 & 10.5 & 5.6 & 11.1
& 16.5 & 13.9 & 9.3 & 14.7
& 19.2 & 15.6 & 10.7 & 16.9 \\

TOUAP~\cite{hupatch}          
& 31.6 & 28.2 & 21.0 & 25.7
& 34.8 & 31.0 & 24.2 & 29.3
& 38.6 & 33.7 & 27.5 & 32.8
& 41.3 & 36.5 & 30.1 & 35.0 \\

AdvPatch~\cite{thys2019fooling} 
& 33.3 & 29.5 & 21.8 & 26.6
& 41.5 & 37.7 & 29.6 & 35.1
& 39.8 & 35.1 & 27.7 & 33.4
& 42.1 & 37.3 & 30.2 & 35.9 \\

UNIPatch~\cite{tpamipatch}    
& 46.9 & 41.6 & 33.7 & 38.8
& 49.2 & 44.9 & 37.5 & 42.6
& 52.0 & 47.3 & 40.3 & 45.2
& 54.5 & 49.4 & 42.8 & 47.7 \\

\rowcolor{gray!20}
\textbf{CDUPatch (Ours)} 
& \textbf{70.4} & \textbf{66.7} & \textbf{57.5} & \textbf{63.2}
& \textbf{71.8} & \textbf{68.5} & \textbf{60.3} & \textbf{65.6}
& \textbf{73.2} & \textbf{73.1} & \textbf{63.0} & \textbf{68.0}
& \textbf{74.6} & \textbf{71.5} & \textbf{65.2} & \textbf{70.7} \\
\bottomrule
\end{tabular}
}
\label{tab:digital exp}
\end{table*}

\begin{table}[t]
\centering
\caption{Black-box attack transferability of patch attacks on the DroneVehicle dataset. Gray diagonals show white-box results; others indicate black-box transfers.}
\begin{tabular}{lcccc}
\toprule
\textbf{Detector} & \textbf{v3} & \textbf{v5} & \textbf{v8} & \textbf{RCNN} \\
\midrule
YOLOv3    & \cellcolor{gray!20}70.4 & 54.0 & 45.2 & 51.5 \\
YOLOv5    & 57.4 & \cellcolor{gray!20}66.7 & 49.0 & 54.7 \\
YOLOv8    & 46.0 & 48.4 & \cellcolor{gray!20}57.5 & 47.5 \\
Fast-RCNN & 52.5 & 54.0 & 46.2 & \cellcolor{gray!20}63.2 \\
\bottomrule
\end{tabular}
\label{black_transfer}
\end{table}

{\flushleft\bf Baselines.}
We compare our method with three kinds of representative patch attacks: 
1) Dual-modal patch-based attacks, which are designed to attack both visible and infrared detectors. We include TOUAP~\cite{hupatch} and UNIPatch~\cite{tpamipatch} as representative approaches that jointly optimize cross-modal adversarial patches. 
2) Single-modal patch-based attacks, which are only optimized in the visible domain. We adopt AdvPatch~\cite{thys2019fooling} as a typical baseline that applies adversarial textures to fool visible-light detectors. 
3) Randomized patch attacks, in which patches with randomly generated shapes, sizes, and colors are placed on the image without any optimization (denoted as RandomPatch). 

{\flushleft\bf Victim Models.}
For the vehicle detection task, we evaluate our attack on two mainstream detection paradigms: one-stage detectors (i.e., YOLOv3~\cite{yolov3}, YOLOv5~\cite{yolov5}, and YOLOv8~\cite{yolov8}) and a two-stage detector (i.e., Faster R-CNN~\cite{fast-rcnn}). For the model training, we choose the officially pre-trained weights as the initialized weights and then fine-tune these models on the training sets of DroneVehicle and MSDrone. After fine-tuning, all detection models achieve over 90\% mean Average Precision (mAP) on the DroneVehicle validation set.

{\flushleft\bf Implementation Details.}
In digital experiments, the adversarial patch is applied to all vehicle targets in each image. It is placed at the top-center of the bounding box and scaled to cover no more than 30\% of the vehicle area. The patch is trained using a combination of the augmented DroneVehicle and MSDrone training sets.
In our physical experiments, we use DJI Matrice 30T~\cite{dji30t} and Matrice 4T~\cite{dji4t} drones equipped with dual-modal sensors to collect real-world visible-infrared data. We print the adversarial patches on foam boards with a fixed width and height of 1.2 meters. The patches are physically deployed on the roof of vehicles, occupying less than 1/3 of the vehicle’s surface area. This placement ensures that critical regions such as the windshield remain unobstructed, thereby maintaining physical feasibility and driver safety.
For hyperparameters in the overall adversarial loss, we set $\gamma$=2.5 and $\delta$=1. 
More implementation details can be found in the Appendix.

\subsection{Digital Domain Performance}

\begin{figure}[t]
\centering
\includegraphics[scale=0.30]{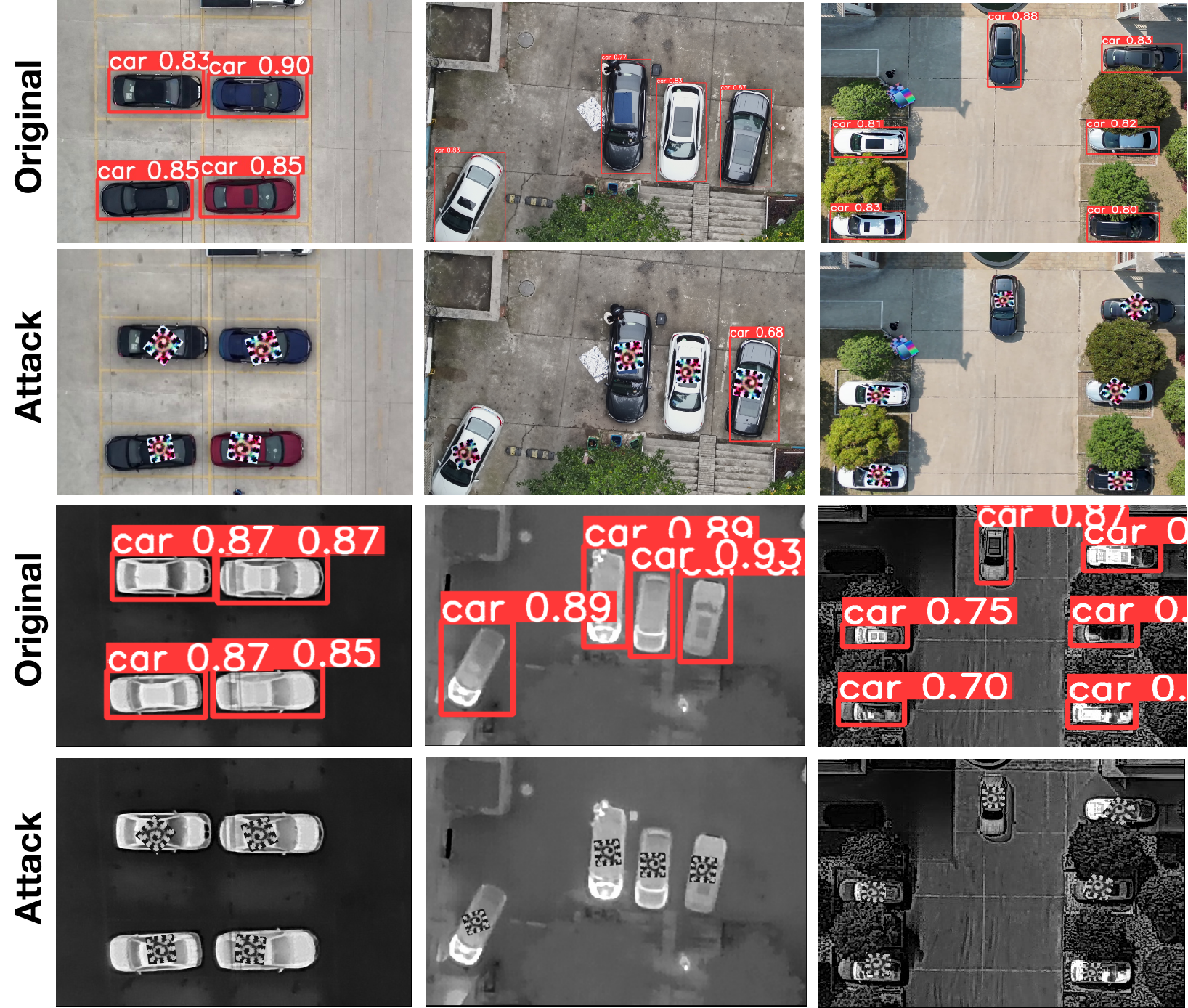}
\caption{Visual examples of our cross-modal patches in the digital world. After our patch attack, the visible-infrared detector fails to accurately recognize the vehicle targets.}
\label{fig:digital exp}
\Description{Placeholder description.}
\end{figure}

{\flushleft\bf Cross-Modal Patch Attack in the Digital World.} Figure~\ref{fig:digital exp} presents visual examples of our cross-modal patches against visible-infrared detection in the digital domain. It demonstrates the attack effectiveness of our patch in the digital domain. Before the attack, the visible-infrared detectors accurately identify the vehicles with high confidence scores. Following the application of our patch, the detectors exhibit a marked degradation in performance, with a significant reduction in detection accuracy and even complete failure to recognize the vehicles. 

{\flushleft\bf Comparing to other patch attacks.} Table~\ref{tab:digital exp} quantitatively compares various adversarial patch attacks on DroneVehicle~\cite{dronevehicle}, VisDrone~\cite{visdrone}, LLVIP~\cite{LLVIP} and MsDrone datasets. As shown, CDUPatch achieves the highest attack success rate across all four datasets
, manifesting its superiority over existing methods (i.e., RandomPatch, TOUAP~\cite{hupatch}, AdvPatch~\cite{thys2019fooling} and UNIPatch~\cite{tpamipatch}, in the context of dual-modal (visible-infrared) vehicle detection. Specifically, we observe that our method maintains strong performance across different dual-modal detectors. For example, on the DroneVehicle dataset, CDUPatch achieves attack success rates of 70.4\% on YOLOv3, 66.7\% on YOLOv5, 57.5\% on YOLOv8, and 63.2\% on Fast-RCNN.
Moreover, CDUPatch exhibits universal effectiveness across different datasets. For instance, when attacking YOLOv5, it achieves ASR of 66.7\%, 68.5\%, 73.1\%, and 71.5\% on DroneVehicle, VisDrone, LLVIP, and MsDrone, respectively. It demonstrates that attack generalizability of CDUPatch across different models and datasets.

\begin{figure*}[t]
\centering
\includegraphics[scale=0.39]{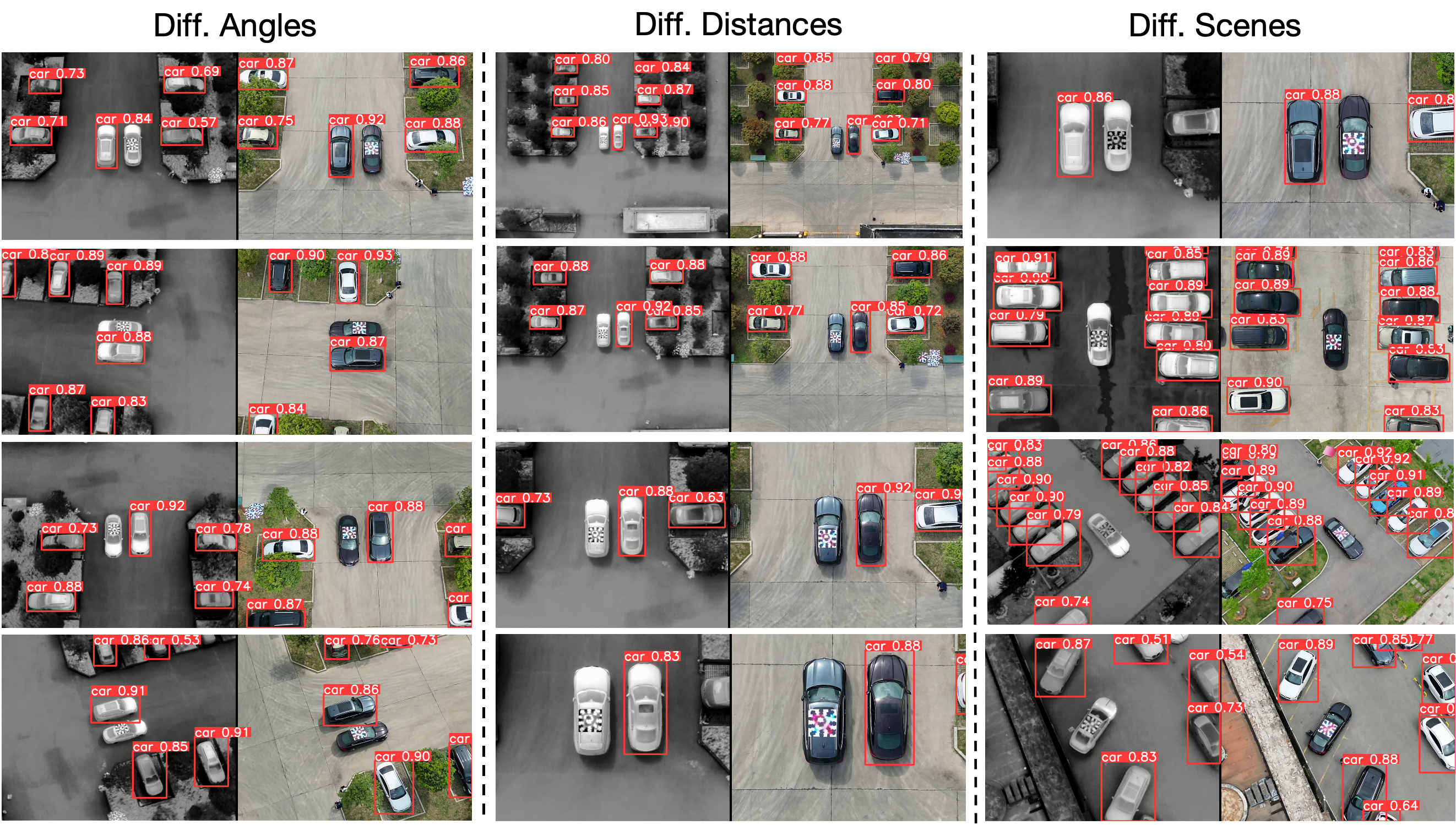}
\vspace{-0.1in}
\caption{Black-box transfer attack of our cross-modal patch in the real world. It demonstrates the attack effectiveness of CDUPatch across different views, distances and scenarios. (refer to Supplementary Materials for a video demo).}
\label{fig:physical_exp}
\Description{Placeholder description.}
\end{figure*}

\begin{figure}[t]
\centering
\includegraphics[scale=0.37]{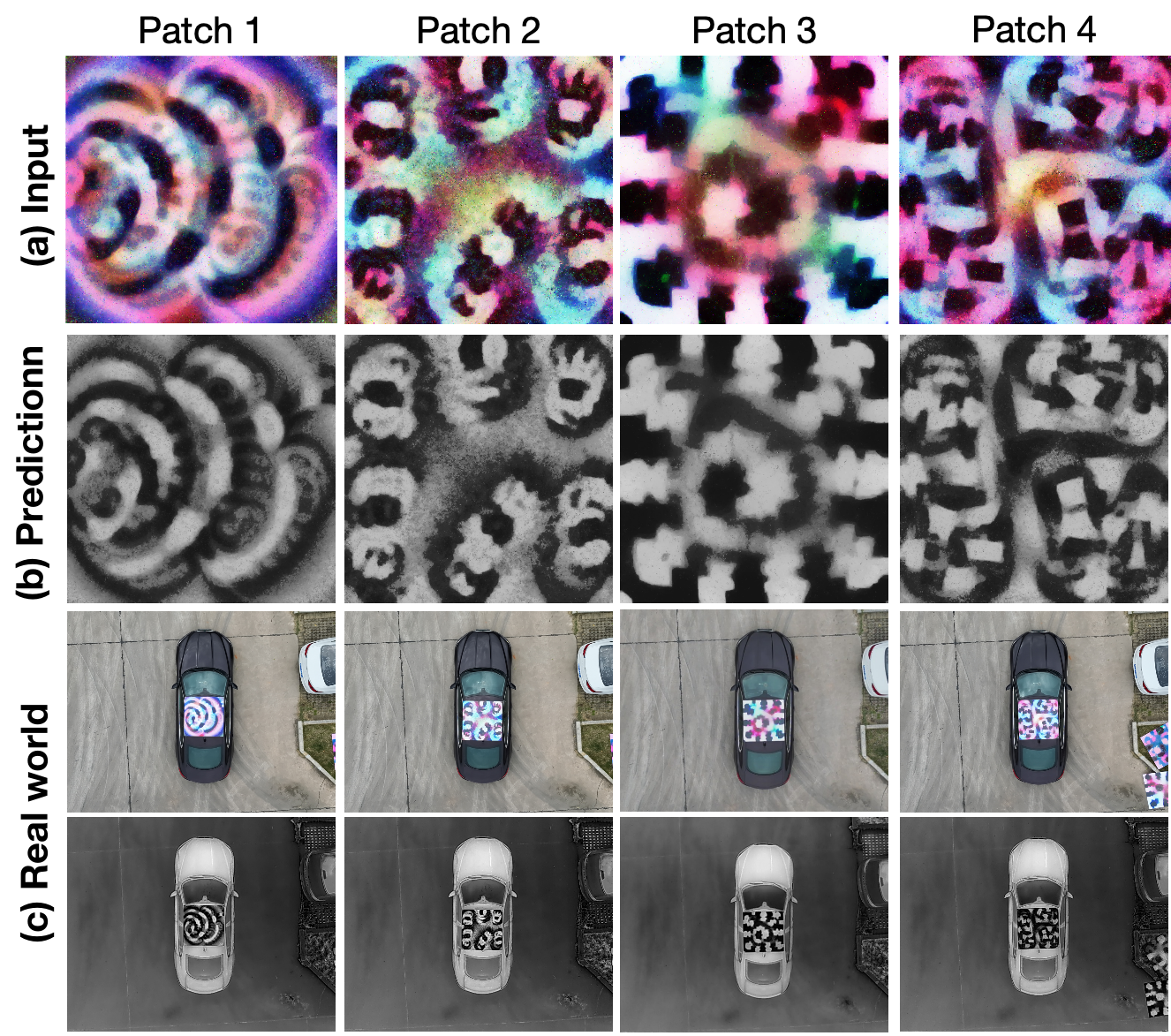}
\vspace{-0.1in}
\caption{Effectiveness validation of our RGB-to-infrared adapter. Top to bottom: the input patches, the predictions of our RGB-to-infrared adapter, and the corresponding visible-infrared imaging captured in the real world. 
}
\label{fig:RGB-infrared validation}
\Description{Placeholder description.}
\end{figure}

{\flushleft\bf Attack Tranferability.} Table~\ref{black_transfer} presents the black-box transferability of patch attacks across different detectors on the DroneVehicle dataset. Diagonal entries (gray) denote white-box attacks, while off-diagonal values indicate black-box transfers across detectors. As shown, the transferred patches still maintain attack effectiveness across black-box models, indicating strong cross-model transferability of our adversarial patches. For example, a patch generated for YOLOv5 performs relatively well when transferred to other models such as YOLOv3 (57.4\%), YOLOv8 (48.4\%), and Fast-RCNN (54.0\%), demonstrating strong transferability across different detectors.

\subsection{Physical Domain Performance}
Although our patches perform well in the digital domain, applying them to the physical world is challenging due to its dynamic nature. In this section, we qualitatively evaluate the effectiveness of our adversarial patches and explore their limitations in different physical environments.

{\flushleft\bf Effectiveness of RGB-to-Infrared adapter.}
Figure~\ref{fig:RGB-infrared validation} shows various patches before and after transformation by our RGB-to-infrared adapter, along with the corresponding infrared images captured in the real world. As shown, the predicted infrared textures closely match the actual infrared observations in the real world, demonstrating that the adapter accurately predicts thermal responses across various color patterns. These results confirm the adapter’s effectiveness in modeling visible-infrared appearance for unified cross-modal patch optimization.

{\flushleft\bf Physical Validation.}
We conduct a black-box transfer attack across various real-world scenarios, as demonstrated in Figure~\ref{fig:physical_exp}. The results highlight the effectiveness of CDUPatch under different angles, distances, and scenes.
In Figure~\ref{fig:physical_exp} (a), an adversarial patch is placed at the top of the car, with different angles captured. The results show that even with substantial angle changes, the attack remains effective under both infrared and visible detection. As the angle changes, the vehicle with the patch consistently goes undetected (e.g., confidence score < 0.5), demonstrating that the patch successfully deceives visible-infrared detectors from various perspectives.
Figure~\ref{fig:physical_exp} (b) presents our attack against scenarios where the distance between the camera and the target vehicles varies (10-80 meters). The effectiveness of the attack remains robust, with the vehicle consistently going undetected as the distance increases, demonstrating the patch's ability to deceive detectors over a range of distances.
Figure~\ref{fig:physical_exp} (c) illustrates detection results on various environmental conditions and scenes. The car with the adversarial patch continues to deceive visible-infrared detectors across diverse natural backgrounds, highlighting the patch's effectiveness in disrupting dual-modal detectors in different real-world environments.

{\flushleft\bf Attack Boundary.}
We examine four key physical conditions that may affect the effectiveness of patch attacks: (a) patch size, (b) distance, (c) viewing angle, and (d) pose angle. Specifically, we distinguish between two types of angles: the viewing angle, which refers to the relative position of the drone camera with respect to the target (e.g., front, side, rear views), and the pose angle, which denotes the heading direction of the vehicle relative to the drone camera (e.g., 0°–90°).  The quantitative
evaluation results of the attack boundary effectiveness for three different patches trained on YOLOv3, YOLOv5, YOLOv8
are presented in Table~\ref{tab:bounds of effectiveness}.

Based on the experimental results, several important conclusions can be drawn. First, the effectiveness of patch attacks varies under different physical conditions. Second, the distance between the patch and the camera must remain within a specific range (10–100 meters) to maintain attack effectiveness. Third, our patch remains effective across a full 360-degree viewing range, including front, side, and rear views. In addition, larger patch sizes generally yield stronger attack performance. However, to avoid occluding the vehicle’s windows, the patch size in CDUPatch is fixed at approximately 30\% of the vehicle’s area (roughly 1.2 $\times$ 1.2m). Lastly, the relative angle between the patch and the camera should ideally remain within a limited range (i.e., 0–50 degrees) to ensure consistent attack success. Overall, our adversarial patch demonstrates robust effectiveness across multi-view and multi-scale conditions within a reasonable physical range.

\begin{table}[t]
\caption{Studies on the effectiveness range of three adversarial patches. 
It reports their effective attack range in terms of viewing distance, angle, patch size, and target pose.
}
\centering
\resizebox{0.48\textwidth}{!}{
\begin{tabular}{ccccc}
\toprule
&   Distance   &   Viewing angle   &   Patch size   &  Pose angle  \\
\midrule
\ Patch 1 & 10m-90m & 0°-360° & 90cm-150cm & \quad 0°-50° \quad \\
\ Patch 2 & 10m-100m & 0°-360° & 80cm-150cm & \quad 0°-60° \quad \\
\ Patch 3 & 20m-90m & 0°-360° & 90cm-150cm & \quad 0°-50° \quad  \\
\bottomrule
\end{tabular}
}
\label{tab:bounds of effectiveness}
\end{table}



\begin{figure}[t]
\centering
\includegraphics[scale=0.26]{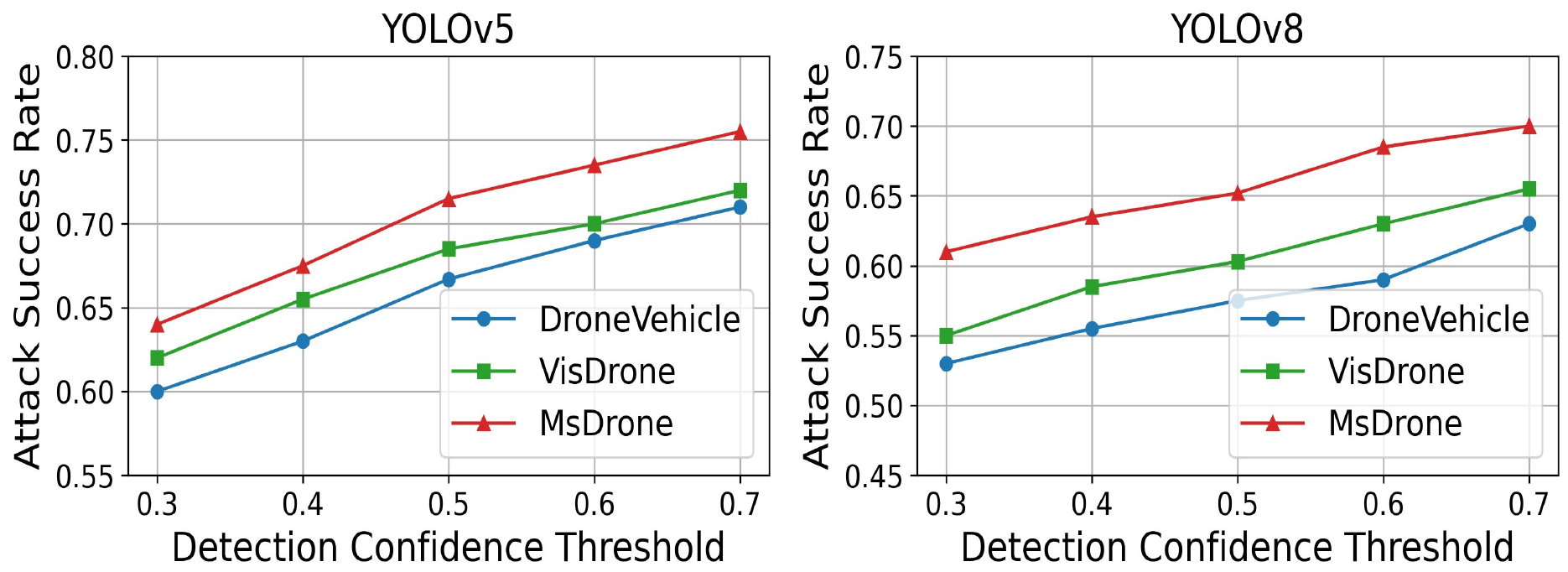}
\caption{ASR under varying detection confidence thresholds on YOLOv5 and YOLOv8. It demonstrates that our patch remains consistently attack effective across a range of threshold settings.}
\label{fig:varying threshold}
\Description{Placeholder description.}
\end{figure}

\begin{table}[t]
\centering
\caption{ASR of different variants of CDUPatch against YOLOv5 and YOLOv8. ``Ada.'' represents the RGB-to-infrared adapter and ``Aug.'' represents our data augmentation strategies.}
\begin{tabular}{ccc}
\toprule
{Victim} & {\quad \quad Attacker} & \multicolumn{1}{c}{ASR (\%)} \\ 
\cmidrule(l){3-3} 
 &  & Visible / Infrared \\
\midrule
 & CDUPatch w/o Ada. & 41.3 / 20.7 \\
 {\ YOLOv5~\cite{yolov5}\ }& CDUPatch w/o Aug. & 55.9 / 59.8 \\
 & CDUPatch & \textbf{69.2 / 73.8} \\ 
\midrule
 & CDUPatch w/o Ada. & 38.5 / 23.3 \\
 YOLOv8~\cite{yolov8}& CDUPatch w/o Aug. & 51.6 / 53.1 \\
 & CDUPatch & \textbf{63.0 / 67.4} \\ 
\bottomrule
\end{tabular}
\label{tab:Effectiveness analysis of the losses on MOT17 dataset}
\end{table}

\begin{figure}[t]
\centering
\begin{tabular}{@{}ccc@{}}
    \toprule
    Patch Color & Visible Attack (\%) & Infrared Attack (\%) \\
    \midrule
    Blue   & 3.5 & 5.2  \\
    Green  & 4.0 & 8.7  \\  
    Red    & 2.9 & 4.6  \\
    White  & 5.5 & 10.9 \\
    Yellow & 3.8 & 7.2  \\
    Ours   & \textbf{71.5} & \textbf{75.6} \\
    \bottomrule
\end{tabular}
\includegraphics[scale=0.25]{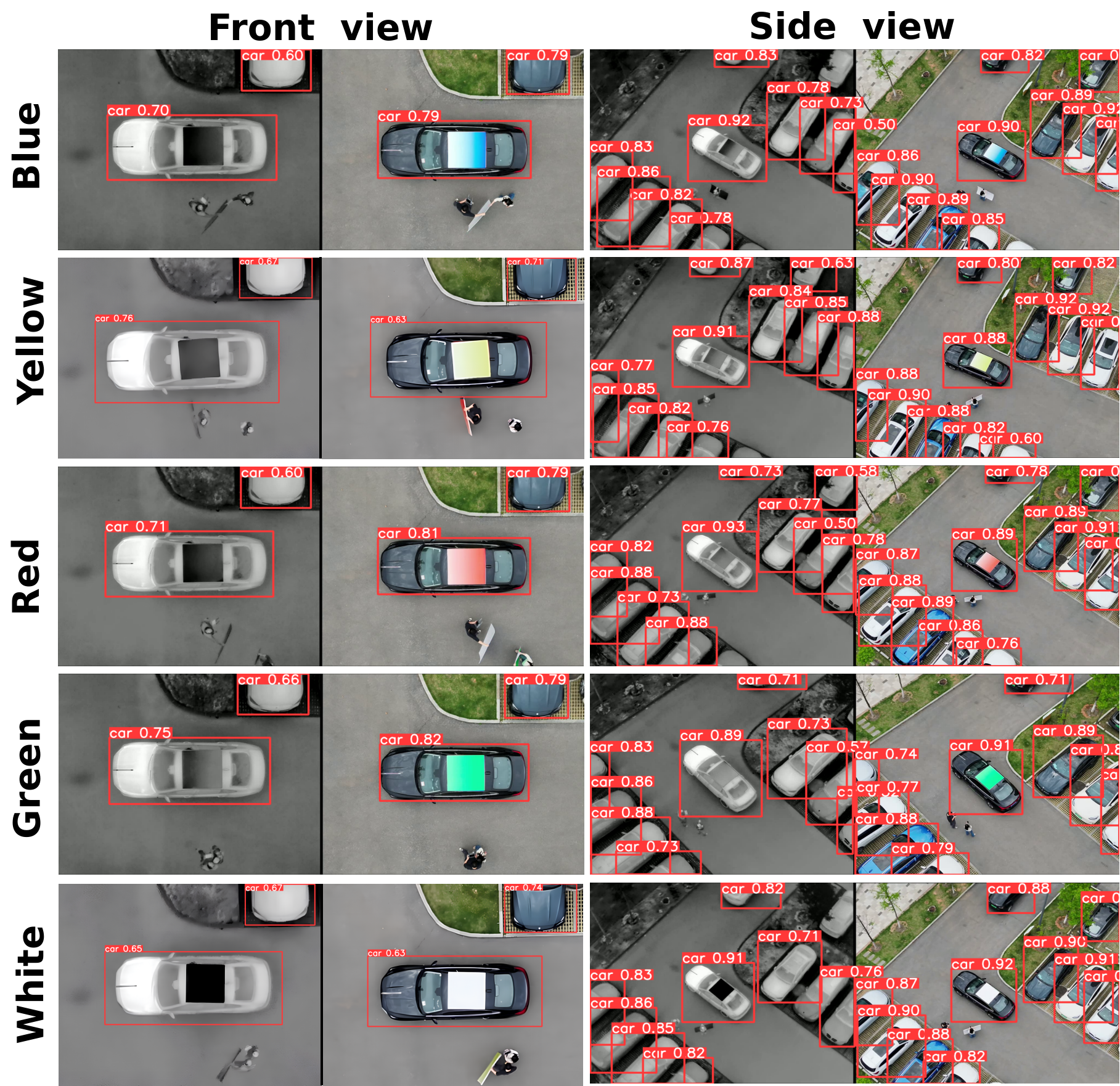}
\caption{Ablation study on the effectiveness of different color patches against visible-infrared detection in the real world. }
\label{fig:ablation_colors}
\Description{Placeholder description.}
\end{figure}

\subsection{Ablation Studies} 
{\flushleft\bf Varying Thresholds.}
We evaluate the sensitivity of our attack to varying detection confidence thresholds. 
As shown in Figure~\ref{fig:varying threshold}, our CDUPatch consistently achieves increasing ASR as the threshold rises from 0.3 to 0.7, indicating stronger attack effectiveness under stricter detection confidences.
For instance, when attacking YOLOv5 on the MSDrone dataset, the ASR remains high at 64.0\%, 67.5\%, 71.4\%, 73.3\%, and 75.6\% under thresholds of 0.3, 0.4, 0.5, 0.6, and 0.7, respectively.
This trend is observed across different detectors (YOLOv5 and YOLOv8) and datasets (DroneVehicle, VisDrone, and MsDrone), 
demonstrating the robustness of our patch against variations in the detection confidence filter. 
In all main experiments, we adopt a default threshold of 0.5, as it offers a good trade-off between detection recall and attack effectiveness.

{\flushleft\bf Varying Colorful Patches.}
We evaluate the effectiveness of attacking the dual-modal detection system using unoptimized patches (e.g., blue, green, red, white and yellow patches). As shown in Figure~\ref{fig:ablation_colors}, despite the application of these colored patches, vehicles remain correctly detected with high confidence across both visible and infrared domains. Moreover, the quantitative results show that their attack success rates remain low, ranging from 5.2\% to 10.9\% under infrared detection, and even lower in the visible detection. In contrast, our optimized patch achieves success rates of 71.5\% and 75.6\% respectively, highlighting the necessity of adversarial optimization to effectively disrupt visible-infrared detectors.

{\flushleft\bf Module Ablations.}
Table~\ref{tab:Effectiveness analysis of the losses on MOT17 dataset} shows how the RGB-to-infrared adapter module (denoted as Ada.) and data augmentation strategies (denoted as Aug.) affect the performance of our CDUPatch against YOLOv5 and YOLOv8 on the MSDrone dataset. 
When attacking YOLOv5, CDUPatch without the RGB-to-Infrared adapter yields significantly lower ASR in both visible and infrared domains (41.3\% / 20.7\%) compared to the full model (69.2\% / 73.8\%), highlighting the critical role of the adapter in disrupting multi-modal detection. 
Similarly, when attacking YOLOv8, removing the augmentation module leads to a considerable ASR drop (from 63.0\% / 67.4\% to 51.6\% / 53.1\%), demonstrating the importance of data augmentation in enhancing patch robustness under varying attack scenarios.



\section{Conclusion}
In this paper, we propose CDUPatch, a color-driven universal adversarial patch framework designed to attack dual-modal object detectors in both visible and infrared modalities. By leveraging the color-thermal relationship observed in real-world materials, we introduce a learnable RGB-to-infrared adapter that maps
RGB patches to simulated infrared patches, which enables joint cross-modal optimization of universal adversarial patches. To enhance attack generalization, we also propose a multi-scale clipping strategy and collect the MSDrone dataset, which features vehicle targets of varying sizes and viewpoints.
Experiments on four benchmark datasets (e.g., DroneVehicle, LLVIP, VisDrone, MSDrone) show that our method outperforms existing patch attacks in the digital domain. Physical-world tests further show CDUPatch has strong transferability across scales, views, and scenarios.


\newpage

\bibliographystyle{ACM-Reference-Format}
\bibliography{main}


\begin{thebibliography}{38}


\ifx \showCODEN    \undefined \def \showCODEN     #1{\unskip}     \fi
\ifx \showISBNx    \undefined \def \showISBNx     #1{\unskip}     \fi
\ifx \showISBNxiii \undefined \def \showISBNxiii  #1{\unskip}     \fi
\ifx \showISSN     \undefined \def \showISSN      #1{\unskip}     \fi
\ifx \showLCCN     \undefined \def \showLCCN      #1{\unskip}     \fi
\ifx \shownote     \undefined \def \shownote      #1{#1}          \fi
\ifx \showarticletitle \undefined \def \showarticletitle #1{#1}   \fi
\ifx \showURL      \undefined \def \showURL       {\relax}        \fi
\providecommand\bibfield[2]{#2}
\providecommand\bibinfo[2]{#2}
\providecommand\natexlab[1]{#1}
\providecommand\showeprint[2][]{arXiv:#2}

\bibitem[30T(2025)]%
        {dji30t}
\bibfield{author}{\bibinfo{person}{DJI~Matrice 30T}.} \bibinfo{year}{2025}\natexlab{}.
\newblock \bibinfo{title}{https://enterprise.dji.com/matrice-30}.
\newblock


\bibitem[4T(2025)]%
        {dji4t}
\bibfield{author}{\bibinfo{person}{DJI~Matrice 4T}.} \bibinfo{year}{2025}\natexlab{}.
\newblock \bibinfo{title}{https://enterprise.dji.com/matrice-4-series}.
\newblock


\bibitem[Athalye et~al\mbox{.}(2018)]%
        {athalye2018synthesizing}
\bibfield{author}{\bibinfo{person}{Anish Athalye}, \bibinfo{person}{Logan Engstrom}, \bibinfo{person}{Andrew Ilyas}, {and} \bibinfo{person}{Kevin Kwok}.} \bibinfo{year}{2018}\natexlab{}.
\newblock \showarticletitle{Synthesizing robust adversarial examples}. In \bibinfo{booktitle}{\emph{International conference on machine learning}}.
\newblock


\bibitem[Brown et~al\mbox{.}(2017)]%
        {brown2017adversarial}
\bibfield{author}{\bibinfo{person}{Tom~B Brown}, \bibinfo{person}{Dandelion Man{\'e}}, \bibinfo{person}{Aurko Roy}, \bibinfo{person}{Mart{\'\i}n Abadi}, {and} \bibinfo{person}{Justin Gilmer}.} \bibinfo{year}{2017}\natexlab{}.
\newblock \showarticletitle{Adversarial patch}.
\newblock \bibinfo{journal}{\emph{arXiv preprint arXiv:1712.09665}} (\bibinfo{year}{2017}).
\newblock


\bibitem[Cheng et~al\mbox{.}(2023)]%
        {cheng2023towards}
\bibfield{author}{\bibinfo{person}{Gong Cheng}, \bibinfo{person}{Xiang Yuan}, \bibinfo{person}{Xiwen Yao}, \bibinfo{person}{Kebing Yan}, \bibinfo{person}{Qinghua Zeng}, \bibinfo{person}{Xingxing Xie}, {and} \bibinfo{person}{Junwei Han}.} \bibinfo{year}{2023}\natexlab{}.
\newblock \showarticletitle{Towards large-scale small object detection: Survey and benchmarks}.
\newblock \bibinfo{journal}{\emph{IEEE Transactions on Pattern Analysis and Machine Intelligence}} (\bibinfo{year}{2023}).
\newblock


\bibitem[Dalal and Triggs(2005)]%
        {inria}
\bibfield{author}{\bibinfo{person}{Navneet Dalal} {and} \bibinfo{person}{Bill Triggs}.} \bibinfo{year}{2005}\natexlab{}.
\newblock \showarticletitle{Histograms of Oriented Gradients for Human Detection}. In \bibinfo{booktitle}{\emph{Proceedings of the IEEE Computer Society Conference on Computer Vision and Pattern Recognition (CVPR)}}. \bibinfo{publisher}{IEEE}.
\newblock


\bibitem[Ding et~al\mbox{.}(2021)]%
        {ding2021object}
\bibfield{author}{\bibinfo{person}{Jian Ding}, \bibinfo{person}{Nan Xue}, \bibinfo{person}{Gui-Song Xia}, \bibinfo{person}{Xiang Bai}, \bibinfo{person}{Wen Yang}, \bibinfo{person}{Michael~Ying Yang}, \bibinfo{person}{Serge Belongie}, \bibinfo{person}{Jiebo Luo}, \bibinfo{person}{Mihai Datcu}, \bibinfo{person}{Marcello Pelillo}, {et~al\mbox{.}}} \bibinfo{year}{2021}\natexlab{}.
\newblock \showarticletitle{Object detection in aerial images: A large-scale benchmark and challenges}.
\newblock \bibinfo{journal}{\emph{IEEE transactions on pattern analysis and machine intelligence}} (\bibinfo{year}{2021}).
\newblock


\bibitem[Girshick(2015)]%
        {fast-rcnn}
\bibfield{author}{\bibinfo{person}{Ross Girshick}.} \bibinfo{year}{2015}\natexlab{}.
\newblock \showarticletitle{Fast r-cnn}. In \bibinfo{booktitle}{\emph{Proceedings of the IEEE international conference on computer vision}}.
\newblock


\bibitem[Hou et~al\mbox{.}(2024)]%
        {hou2024object}
\bibfield{author}{\bibinfo{person}{Zhiqiang Hou}, \bibinfo{person}{Chen Yang}, \bibinfo{person}{Ying Sun}, \bibinfo{person}{Sugang Ma}, \bibinfo{person}{Xiaobao Yang}, {and} \bibinfo{person}{Jiulun Fan}.} \bibinfo{year}{2024}\natexlab{}.
\newblock \showarticletitle{An object detection algorithm based on infrared-visible dual modal feature fusion}.
\newblock \bibinfo{journal}{\emph{Infrared Physics \& Technology}} (\bibinfo{year}{2024}).
\newblock


\bibitem[Howell et~al\mbox{.}(2020)]%
        {howell2020thermal}
\bibfield{author}{\bibinfo{person}{John~R Howell}, \bibinfo{person}{M~Pinar Meng{\"u}{\c{c}}}, \bibinfo{person}{Kyle Daun}, {and} \bibinfo{person}{Robert Siegel}.} \bibinfo{year}{2020}\natexlab{}.
\newblock \bibinfo{booktitle}{\emph{Thermal radiation heat transfer}}.
\newblock \bibinfo{publisher}{CRC press}.
\newblock


\bibitem[Hu et~al\mbox{.}(2025)]%
        {hupatch}
\bibfield{author}{\bibinfo{person}{Chengyin Hu}, \bibinfo{person}{Weiwen Shi}, \bibinfo{person}{Wen Yao}, \bibinfo{person}{Tingsong Jiang}, \bibinfo{person}{Ling Tian}, {and} \bibinfo{person}{Wen Li}.} \bibinfo{year}{2025}\natexlab{}.
\newblock \showarticletitle{Two-stage optimized unified adversarial patch for attacking visible-infrared cross-modal detectors in the physical world}.
\newblock \bibinfo{journal}{\emph{Applied Soft Computing}} (\bibinfo{year}{2025}).
\newblock


\bibitem[Hu et~al\mbox{.}(2021)]%
        {hu2021naturalistic}
\bibfield{author}{\bibinfo{person}{Yu-Chih-Tuan Hu}, \bibinfo{person}{Bo-Han Kung}, \bibinfo{person}{Daniel~Stanley Tan}, \bibinfo{person}{Jun-Cheng Chen}, \bibinfo{person}{Kai-Lung Hua}, {and} \bibinfo{person}{Wen-Huang Cheng}.} \bibinfo{year}{2021}\natexlab{}.
\newblock \showarticletitle{Naturalistic physical adversarial patch for object detectors}. In \bibinfo{booktitle}{\emph{Proceedings of the IEEE/CVF International Conference on Computer Vision}}.
\newblock


\bibitem[Hu et~al\mbox{.}(2022)]%
        {hu2022adversarial}
\bibfield{author}{\bibinfo{person}{Zhanhao Hu}, \bibinfo{person}{Siyuan Huang}, \bibinfo{person}{Xiaopei Zhu}, \bibinfo{person}{Fuchun Sun}, \bibinfo{person}{Bo Zhang}, {and} \bibinfo{person}{Xiaolin Hu}.} \bibinfo{year}{2022}\natexlab{}.
\newblock \showarticletitle{Adversarial texture for fooling person detectors in the physical world}. In \bibinfo{booktitle}{\emph{Proceedings of the IEEE/CVF conference on computer vision and pattern recognition}}.
\newblock


\bibitem[Huang et~al\mbox{.}(2023)]%
        {TSEA}
\bibfield{author}{\bibinfo{person}{Hao Huang}, \bibinfo{person}{Ziyan Chen}, \bibinfo{person}{Huanran Chen}, \bibinfo{person}{Yongtao Wang}, {and} \bibinfo{person}{Kevin Zhang}.} \bibinfo{year}{2023}\natexlab{}.
\newblock \showarticletitle{T-sea: Transfer-based self-ensemble attack on object detection}. In \bibinfo{booktitle}{\emph{Proceedings of the IEEE/CVF conference on computer vision and pattern recognition}}.
\newblock


\bibitem[Jia et~al\mbox{.}(2020)]%
        {jia2020robust}
\bibfield{author}{\bibinfo{person}{Shuai Jia}, \bibinfo{person}{Chao Ma}, \bibinfo{person}{Yibing Song}, {and} \bibinfo{person}{Xiaokang Yang}.} \bibinfo{year}{2020}\natexlab{}.
\newblock \showarticletitle{Robust tracking against adversarial attacks}. In \bibinfo{booktitle}{\emph{European Conference on Computer Vision}}.
\newblock


\bibitem[Jia et~al\mbox{.}(2022)]%
        {jia2022exploring}
\bibfield{author}{\bibinfo{person}{Shuai Jia}, \bibinfo{person}{Chao Ma}, \bibinfo{person}{Taiping Yao}, \bibinfo{person}{Bangjie Yin}, \bibinfo{person}{Shouhong Ding}, {and} \bibinfo{person}{Xiaokang Yang}.} \bibinfo{year}{2022}\natexlab{}.
\newblock \showarticletitle{Exploring frequency adversarial attacks for face forgery detection}. In \bibinfo{booktitle}{\emph{Proceedings of the IEEE/CVF Conference on Computer Vision and Pattern Recognition}}.
\newblock


\bibitem[Jia et~al\mbox{.}(2021a)]%
        {jia2021iou}
\bibfield{author}{\bibinfo{person}{Shuai Jia}, \bibinfo{person}{Yibing Song}, \bibinfo{person}{Chao Ma}, {and} \bibinfo{person}{Xiaokang Yang}.} \bibinfo{year}{2021}\natexlab{a}.
\newblock \showarticletitle{Iou attack: Towards temporally coherent black-box adversarial attack for visual object tracking}. In \bibinfo{booktitle}{\emph{Proceedings of the IEEE/CVF Conference on Computer Vision and Pattern Recognition}}.
\newblock


\bibitem[Jia et~al\mbox{.}(2021b)]%
        {LLVIP}
\bibfield{author}{\bibinfo{person}{Xinyu Jia}, \bibinfo{person}{Chuang Zhu}, \bibinfo{person}{Minzhen Li}, \bibinfo{person}{Wenqi Tang}, {and} \bibinfo{person}{Wenli Zhou}.} \bibinfo{year}{2021}\natexlab{b}.
\newblock \showarticletitle{LLVIP: A visible-infrared paired dataset for low-light vision}. In \bibinfo{booktitle}{\emph{Proceedings of the IEEE/CVF international conference on computer vision}}.
\newblock


\bibitem[Jocher(2020)]%
        {yolov5}
\bibfield{author}{\bibinfo{person}{Glenn Jocher}.} \bibinfo{year}{2020}\natexlab{}.
\newblock \bibinfo{booktitle}{\emph{{YOLOv5 by Ultralytics}}}.
\newblock
\href{https://doi.org/10.5281/zenodo.3908559}{doi:\nolinkurl{10.5281/zenodo.3908559}}


\bibitem[Jocher et~al\mbox{.}(2023)]%
        {yolov8}
\bibfield{author}{\bibinfo{person}{Glenn Jocher}, \bibinfo{person}{Ayush Chaurasia}, {and} \bibinfo{person}{Jing Qiu}.} \bibinfo{year}{2023}\natexlab{}.
\newblock \bibinfo{booktitle}{\emph{{YOLO by Ultralytics}}}.
\newblock
\urldef\tempurl%
\url{https://github.com/ultralytics/ultralytics}
\showURL{%
\tempurl}


\bibitem[Liu et~al\mbox{.}(2018)]%
        {liu2018dpatch}
\bibfield{author}{\bibinfo{person}{Xin Liu}, \bibinfo{person}{Huanrui Yang}, \bibinfo{person}{Ziwei Liu}, \bibinfo{person}{Linghao Song}, \bibinfo{person}{Hai Li}, {and} \bibinfo{person}{Yiran Chen}.} \bibinfo{year}{2018}\natexlab{}.
\newblock \showarticletitle{Dpatch: An adversarial patch attack on object detectors}.
\newblock \bibinfo{journal}{\emph{arXiv preprint arXiv:1806.02299}} (\bibinfo{year}{2018}).
\newblock


\bibitem[Long et~al\mbox{.}(2024)]%
        {long2024papmot}
\bibfield{author}{\bibinfo{person}{Jiahuan Long}, \bibinfo{person}{Tingsong Jiang}, \bibinfo{person}{Wen Yao}, \bibinfo{person}{Shuai Jia}, \bibinfo{person}{Weijia Zhang}, \bibinfo{person}{Weien Zhou}, \bibinfo{person}{Chao Ma}, {and} \bibinfo{person}{Xiaoqian Chen}.} \bibinfo{year}{2024}\natexlab{}.
\newblock \showarticletitle{PapMOT: Exploring Adversarial Patch Attack Against Multiple Object Tracking}. In \bibinfo{booktitle}{\emph{European Conference on Computer Vision}}.
\newblock


\bibitem[Long et~al\mbox{.}(2025)]%
        {long2025robust}
\bibfield{author}{\bibinfo{person}{Jiahuan Long}, \bibinfo{person}{Zhengqin Xu}, \bibinfo{person}{Tingsong Jiang}, \bibinfo{person}{Wen Yao}, \bibinfo{person}{Shuai Jia}, \bibinfo{person}{Chao Ma}, {and} \bibinfo{person}{Xiaoqian Chen}.} \bibinfo{year}{2025}\natexlab{}.
\newblock \showarticletitle{Robust SAM: On the Adversarial Robustness of Vision Foundation Models}. In \bibinfo{booktitle}{\emph{Proceedings of the AAAI Conference on Artificial Intelligence}}.
\newblock


\bibitem[Redmon and Farhadi(2018)]%
        {yolov3}
\bibfield{author}{\bibinfo{person}{Joseph Redmon} {and} \bibinfo{person}{Ali Farhadi}.} \bibinfo{year}{2018}\natexlab{}.
\newblock \showarticletitle{Yolov3: An incremental improvement}.
\newblock \bibinfo{journal}{\emph{arXiv preprint arXiv:1804.02767}} (\bibinfo{year}{2018}).
\newblock


\bibitem[Sharif et~al\mbox{.}(2016)]%
        {Sharif_Bhagavatula_Bauer_Reiter_2016}
\bibfield{author}{\bibinfo{person}{Mahmood Sharif}, \bibinfo{person}{Sruti Bhagavatula}, \bibinfo{person}{Lujo Bauer}, {and} \bibinfo{person}{Michael~K. Reiter}.} \bibinfo{year}{2016}\natexlab{}.
\newblock \showarticletitle{Accessorize to a Crime}. In \bibinfo{booktitle}{\emph{Proceedings of the 2016 ACM SIGSAC Conference on Computer and Communications Security}}.
\newblock


\bibitem[Sun et~al\mbox{.}(2022)]%
        {dronevehicle}
\bibfield{author}{\bibinfo{person}{Yiming Sun}, \bibinfo{person}{Bing Cao}, \bibinfo{person}{Pengfei Zhu}, {and} \bibinfo{person}{Qinghua Hu}.} \bibinfo{year}{2022}\natexlab{}.
\newblock \showarticletitle{Drone-based RGB-Infrared Cross-Modality Vehicle Detection via Uncertainty-Aware Learning}.
\newblock \bibinfo{journal}{\emph{IEEE Transactions on Circuits and Systems for Video Technology}} (\bibinfo{year}{2022}).
\newblock
\href{https://doi.org/10.1109/TCSVT.2022.3168279}{doi:\nolinkurl{10.1109/TCSVT.2022.3168279}}


\bibitem[Tang et~al\mbox{.}(2023)]%
        {tang2023adversarial}
\bibfield{author}{\bibinfo{person}{Guijian Tang}, \bibinfo{person}{Tingsong Jiang}, \bibinfo{person}{Weien Zhou}, \bibinfo{person}{Chao Li}, \bibinfo{person}{Wen Yao}, {and} \bibinfo{person}{Yong Zhao}.} \bibinfo{year}{2023}\natexlab{}.
\newblock \showarticletitle{Adversarial patch attacks against aerial imagery object detectors}.
\newblock \bibinfo{journal}{\emph{Neurocomputing}} (\bibinfo{year}{2023}).
\newblock


\bibitem[Thys et~al\mbox{.}(2019)]%
        {thys2019fooling}
\bibfield{author}{\bibinfo{person}{Simen Thys}, \bibinfo{person}{Wiebe Van~Ranst}, {and} \bibinfo{person}{Toon Goedem{\'e}}.} \bibinfo{year}{2019}\natexlab{}.
\newblock \showarticletitle{Fooling automated surveillance cameras: adversarial patches to attack person detection}. In \bibinfo{booktitle}{\emph{Proceedings of the IEEE/CVF conference on computer vision and pattern recognition workshops}}.
\newblock


\bibitem[Wang et~al\mbox{.}(2022)]%
        {FCA}
\bibfield{author}{\bibinfo{person}{Donghua Wang}, \bibinfo{person}{Tingsong Jiang}, \bibinfo{person}{Jialiang Sun}, \bibinfo{person}{Weien Zhou}, \bibinfo{person}{Zhiqiang Gong}, \bibinfo{person}{Xiaoya Zhang}, \bibinfo{person}{Wen Yao}, {and} \bibinfo{person}{Xiaoqian Chen}.} \bibinfo{year}{2022}\natexlab{}.
\newblock \showarticletitle{Fca: Learning a 3d full-coverage vehicle camouflage for multi-view physical adversarial attack}. In \bibinfo{booktitle}{\emph{Proceedings of the AAAI conference on artificial intelligence}}.
\newblock


\bibitem[Wang et~al\mbox{.}(2023)]%
        {wang2023interactively}
\bibfield{author}{\bibinfo{person}{Di Wang}, \bibinfo{person}{Jinyuan Liu}, \bibinfo{person}{Risheng Liu}, {and} \bibinfo{person}{Xin Fan}.} \bibinfo{year}{2023}\natexlab{}.
\newblock \showarticletitle{An interactively reinforced paradigm for joint infrared-visible image fusion and saliency object detection}.
\newblock \bibinfo{journal}{\emph{Information Fusion}} (\bibinfo{year}{2023}).
\newblock


\bibitem[Wei et~al\mbox{.}(2023a)]%
        {wei2023iccv}
\bibfield{author}{\bibinfo{person}{Xingxing Wei}, \bibinfo{person}{Yao Huang}, \bibinfo{person}{Yitong Sun}, {and} \bibinfo{person}{Jie Yu}.} \bibinfo{year}{2023}\natexlab{a}.
\newblock \showarticletitle{Unified Adversarial Patch for Cross-modal Attacks in the Physical World}. In \bibinfo{booktitle}{\emph{Proceedings of the IEEE/CVF International Conference on Computer Vision (ICCV)}}.
\newblock


\bibitem[Wei et~al\mbox{.}(2023b)]%
        {tpamipatch}
\bibfield{author}{\bibinfo{person}{Xingxing Wei}, \bibinfo{person}{Yao Huang}, \bibinfo{person}{Yitong Sun}, {and} \bibinfo{person}{Jie Yu}.} \bibinfo{year}{2023}\natexlab{b}.
\newblock \showarticletitle{Unified adversarial patch for visible-infrared cross-modal attacks in the physical world}.
\newblock \bibinfo{journal}{\emph{IEEE Transactions on Pattern Analysis and Machine Intelligence}} (\bibinfo{year}{2023}).
\newblock


\bibitem[Wei et~al\mbox{.}(2023c)]%
        {weiinfra}
\bibfield{author}{\bibinfo{person}{Xingxing Wei}, \bibinfo{person}{Jie Yu}, {and} \bibinfo{person}{Yao Huang}.} \bibinfo{year}{2023}\natexlab{c}.
\newblock \showarticletitle{Physically adversarial infrared patches with learnable shapes and locations}. In \bibinfo{booktitle}{\emph{Proceedings of the IEEE/CVF conference on computer vision and pattern recognition}}.
\newblock


\bibitem[Wu et~al\mbox{.}(2025)]%
        {wu2025gradient}
\bibfield{author}{\bibinfo{person}{Junqi Wu}, \bibinfo{person}{Wen Yao}, \bibinfo{person}{Shuai Jia}, \bibinfo{person}{Tingsong Jiang}, \bibinfo{person}{Weien Zhou}, \bibinfo{person}{Chao Ma}, {and} \bibinfo{person}{Xiaoqian Chen}.} \bibinfo{year}{2025}\natexlab{}.
\newblock \showarticletitle{Gradient-based sparse voxel attacks on point cloud object detection}.
\newblock \bibinfo{journal}{\emph{Pattern Recognition}} (\bibinfo{year}{2025}).
\newblock


\bibitem[Xu et~al\mbox{.}(2020)]%
        {xu2020adversarial1}
\bibfield{author}{\bibinfo{person}{Han Xu}, \bibinfo{person}{Yao Ma}, \bibinfo{person}{Hao-Chen Liu}, \bibinfo{person}{Debayan Deb}, \bibinfo{person}{Hui Liu}, \bibinfo{person}{Ji-Liang Tang}, {and} \bibinfo{person}{Anil~K Jain}.} \bibinfo{year}{2020}\natexlab{}.
\newblock \showarticletitle{Adversarial attacks and defenses in images, graphs and text: A review}.
\newblock \bibinfo{journal}{\emph{International Journal of Automation and Computing}} (\bibinfo{year}{2020}).
\newblock


\bibitem[Zhang et~al\mbox{.}(2023)]%
        {zhang2023differential}
\bibfield{author}{\bibinfo{person}{Ruiheng Zhang}, \bibinfo{person}{Lu Li}, \bibinfo{person}{Qi Zhang}, \bibinfo{person}{Jin Zhang}, \bibinfo{person}{Lixin Xu}, \bibinfo{person}{Baomin Zhang}, {and} \bibinfo{person}{Binglu Wang}.} \bibinfo{year}{2023}\natexlab{}.
\newblock \showarticletitle{Differential feature awareness network within antagonistic learning for infrared-visible object detection}.
\newblock \bibinfo{journal}{\emph{IEEE Transactions on Circuits and Systems for Video Technology}} (\bibinfo{year}{2023}).
\newblock


\bibitem[Zhu et~al\mbox{.}(2021)]%
        {visdrone}
\bibfield{author}{\bibinfo{person}{Pengfei Zhu}, \bibinfo{person}{Longyin Wen}, \bibinfo{person}{Dawei Du}, \bibinfo{person}{Xiao Bian}, \bibinfo{person}{Heng Fan}, \bibinfo{person}{Qinghua Hu}, {and} \bibinfo{person}{Haibin Ling}.} \bibinfo{year}{2021}\natexlab{}.
\newblock \showarticletitle{Detection and tracking meet drones challenge}.
\newblock \bibinfo{journal}{\emph{IEEE Transactions on Pattern Analysis and Machine Intelligence}} (\bibinfo{year}{2021}).
\newblock


\bibitem[Zhu et~al\mbox{.}(2023)]%
        {zhu2023tpatch}
\bibfield{author}{\bibinfo{person}{Wenjun Zhu}, \bibinfo{person}{Xiaoyu Ji}, \bibinfo{person}{Yushi Cheng}, \bibinfo{person}{Shibo Zhang}, {and} \bibinfo{person}{Wenyuan Xu}.} \bibinfo{year}{2023}\natexlab{}.
\newblock \showarticletitle{$\{$TPatch$\}$: A triggered physical adversarial patch}. In \bibinfo{booktitle}{\emph{32nd USENIX Security Symposium}}.
\newblock


\end{thebibliography}

\end{document}